\documentclass[letterpaper, 10 pt, conference]{ieeeconf}  

\IEEEoverridecommandlockouts                              

\usepackage{graphics} 
\usepackage{epsfig} 
\usepackage{mathptmx} 
\usepackage{times} 
\usepackage{textcomp} 
\usepackage{amsmath} 
\usepackage{amssymb}  
\usepackage{booktabs}

\title{\LARGE \bf
$\pi_0$-EqM: Equilibrium Matching for Closed-Loop Vision-Language-Action Control
}

\author{
Huanming Liu$^{1,*}$, Congsheng Xu$^{2,*}$, Jianmin Ji$^{1,\dagger}$, and Yao Mu$^{2,\dagger}$%
\thanks{$^{1}$Huanming Liu and Jianmin Ji are with University of Science and Technology of China, China
        {\tt\small lhm\_0410@mail.ustc.edu.cn, jianmin@ustc.edu.cn}}%
\thanks{$^{2}$Congsheng Xu and Yao Mu are with Shanghai Jiao Tong University, China
        {\tt\small acondaway@sjtu.edu.cn, muyao@sjtu.edu.cn}}%
\thanks{$^{*}$Equal contribution.}%
\thanks{$^{\dagger}$Corresponding authors.}%
}

\begin{document}

\maketitle
\thispagestyle{empty}
\pagestyle{empty}

\begin{abstract}
Currently, Vision-Language-Action (VLA) models have become the most adopted paradigm for robotic manipulation for its great potential for task generalization. While most generative flow-matching action decoders for VLA control are often deployed with fixed sampling horizons, limiting state-dependent compute and temporal reuse across control cycles. We present $\pi_0$-EqM, which replaces the flow-matching expert in $\pi_0$ with an Equilibrium Matching (EqM) decoder while leaving the upstream VLA stack unchanged. Under a matched 300-step budget, $\pi_0$-EqM improves RoboTwin average success from 40.4\% to 50.2\% across 19 tasks and remains competitive on LIBERO, with its clearest gain on LIBERO-10 (87.0\%). Two threshold scans reveal a task-dependent \textit{non-monotonic} relation between residual and success, which we term the \textit{stationarity--executability gap}. The results suggest that inference depth in iterative VLA control is part of policy design and introduce an energy-based VLA perspective that may inform future work on composable action generation across tasks and embodiments.
\end{abstract}

\section{Introduction}
Vision-Language-Action (VLA) policies map visual observations and language instructions directly to actions \cite{brohan2022rt, zitkovich2023rt, team2024octo, kim2024openvla}. In these systems, a central design choice is the action decoder that produces short action chunks for receding-horizon execution. Recent large-scale VLAs such as $\pi_0$ use flow-matching decoders \cite{black2024pi_0, lipman2022flow}, while diffusion-based action generation remains a strong alternative \cite{chi2025diffusion}. These decoders are expressive, but are typically deployed with fixed denoising or integration horizons.

\begin{figure}[t]
    \centering
    \includegraphics[width=\linewidth]{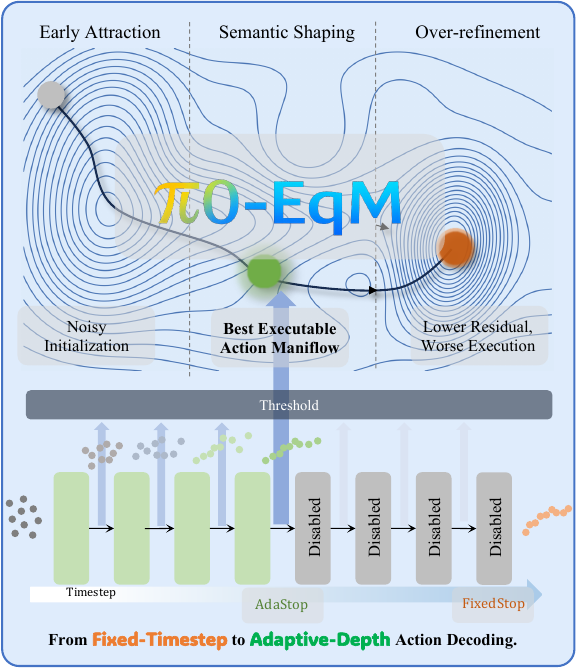}
    \vspace{-2em}
    \caption{Overview of $\pi_0$-EqM. We replace only the action decoder in $\pi_0$ and cast action generation as iterative equilibrium solving, enabling adaptive stopping and warm starts. An executable intermediate action may appear before full numerical convergence.}
    \label{fig:teaser}
    \vspace{-1em}
\end{figure}

This is restrictive in closed loop. Different states benefit from different amounts of computation, and consecutive control cycles are strongly correlated, making iterate reuse desirable. Yet in diffusion and flow decoders, intermediate states are tied to explicit time or noise semantics, which makes early stopping and cross-cycle warm starts awkward.

We therefore ask whether inference depth should be treated as part of deployed policy design. To study this, we replace the flow-matching expert in $\pi_0$ with an Equilibrium Matching (EqM) decoder \cite{wang2025equilibrium}. EqM learns a time-free conditional vector field and decodes actions by iterative equilibrium solving rather than time-indexed transport.

This also introduces an energy-based perspective into VLA and points to composable energy-based action generation across tasks and embodiments as a natural extension. Because EqM is time free, intermediate iterates no longer carry explicit diffusion-time semantics, making stopping and reuse deployment decisions rather than sampler-specific interventions.

Under a matched 300-step inference budget, $\pi_0$-EqM improves RoboTwin average success from 40.4\% to 50.2\% across 19 tasks. On LIBERO, it remains competitive overall and improves LIBERO-10 from 85.2\% to 87.0\%. Threshold scans further show that lower residual does not uniformly imply better execution. We refer to this mismatch as the \emph{stationarity--executability gap}.

Our contributions are summarized as follows:
\begin{itemize}
    \item We introduce $\pi_0$-EqM, a decoder replacement for $\pi_0$ that improves RoboTwin average success by 9.8 points under a matched 300-step budget while remaining competitive on LIBERO.
    \item We show that stopping depth can change closed-loop behavior in EqM-based control.
    \item We connect a local solver analysis to threshold-scan behavior and use the stationarity--executability gap to interpret iterative action decoding.
    \item Our method qualitatively and quantitatively shows that the time-independent diffusion action generation method can lead to better performance for robotic manipulation.
\end{itemize}

\section{Related Work}
\paragraph{VLA policies and chunked action decoding}
Modern VLA systems combine visual encoders, language conditioning, and chunked action prediction in a single model interface~\cite{brohan2022rt,zitkovich2023rt,team2024octo,kim2024openvla}. 
Action chunking improves stability in long-horizon control~\cite{zhao2023learning} and mitigate the non-marcov property of the robotic manipulation. Usually, VLAs utilizes a web-scale data pre-trained VLM as the understanding backbone and discrete action token output, and a light-weight head for continuous action generation based on the VLM's visual and action semantic information. $\pi$ series models extends this line with a flow-matching action expert~\cite{black2024pi_0}; 
our work keeps the structure of upstream VLM and modifies only the action decoder that follows the design art of Equilibrium Matching.


\paragraph{Optimization-based control and energy-based inference}
A complementary line treats action generation as optimization, either by solving a control objective online or by minimizing a learned conditional energy / following its gradient field~\cite{tassa2014control,amos2018differentiable,du2019implicit,florence2022implicit}. Compared with fixed-horizon generative decoding, such formulations make the iterative structure of inference explicit and render test-time computation more controllable, which is particularly relevant in closed-loop settings. EqM is related to this family in its inference form, but differs in parameterization: instead of requiring an explicit scalar energy, it learns an implicit conditional vector field whose equilibria define the target action chunks~\cite{wang2025equilibrium}. Recent work has also explored time-unconditional flow matching for adaptive robotic control~\cite{zhang2026generative}. This makes EqM a natural fit for exposing adaptive stopping and warm-start reuse within a VLA decoder.

\section{Method}

\subsection{Preliminaries: Action Decoding as Iterative Inference}
At control step $t$, a VLA policy maps the multimodal condition $c_t = (o_t, \ell, s_t)$ to an action chunk $A_t = [a_t, a_{t+1}, \dots, a_{t+H-1}] \in \mathbb{R}^{H \times d}$. In our receding-horizon setup, only the immediate action $a_t$ is executed before replanning.

Rather than treating decoding as a fixed-cost black box, we view it as iterative inference whose depth is a controllable policy variable. EqM provides the mathematical scaffolding for this view.

\subsection{Equilibrium Matching for VLA Control}
Unlike diffusion or flow-matching policies that parameterize time-indexed score or velocity fields, EqM learns a \textbf{time-invariant} (autonomous) conditional vector field $f_\theta(A; c_t) \in \mathbb{R}^{H \times d}$. The roots of this field correspond to the target equilibrium action chunks.

\paragraph{Training Objective} Given a demonstration chunk $A \sim p_{\text{data}}$, structural noise $\epsilon \sim \mathcal{N}(0, I)$, and an interpolation factor $\gamma \sim \mathcal{U}(0,1)$, we define the interpolant $A_\gamma = \gamma A + (1-\gamma)\epsilon$. The field $f_\theta$ is trained to match a targeted vector $v_{\text{target}}(\gamma; A, \epsilon) = w(\gamma)(A-\epsilon)$ via the EqM loss:
\begin{equation}
    \mathcal{L}_{\text{EqM}}(\theta) = \mathbb{E}_{\gamma,A,\epsilon} \left[ \left\| f_\theta(A_\gamma; c_t) - w(\gamma)(A-\epsilon) \right\|_2^2 \right],
    \label{eq:eqm_loss}
\end{equation}
where $w(\gamma)$ is a scheduling scalar designed to decay to zero as $\gamma \rightarrow 1$. This enforces the data manifold to act as a set of stable equilibria.

\paragraph{Inference as Equilibrium Solving} At test time, actions are recovered by solving $f_\theta(A; c_t) = 0$. Starting from $A_t^{(0)}$ and setting $A_t^{(-1)} = A_t^{(0)}$, we use a Nesterov-accelerated solver:
\begin{align}
    \widetilde{A}_t^{(k)} &= A_t^{(k)} + \mu \bigl(A_t^{(k)} - A_t^{(k-1)}\bigr), \label{eq:nag_lookahead} \\
    A_t^{(k+1)} &= \widetilde{A}_t^{(k)} - \eta f_\theta(\widetilde{A}_t^{(k)}; c_t), \label{eq:nag}
\end{align}
where $\mu \in [0, 1)$ is the momentum factor and $\eta > 0$ is the step size. We use the same lookahead point $\widetilde{A}_t^{(k)}$ for both the update and the residual check.

\subsection{Temporal Coherence: Adaptive Stopping and Warm-Start}
The iterative, time-invariant nature of EqM exposes two deployment mechanisms:

\textbf{Adaptive Inference Depth:} We define the normalized lookahead residual
\begin{equation}
    r_k = \frac{\left\| f_\theta\bigl(\widetilde{A}_t^{(k)}; c_t\bigr) \right\|_2}{\sqrt{Hd}},
    \label{eq:residual}
\end{equation}
and stop at $T_t(\tau) = \min \{k \ge 0 : r_k \le \tau \text{ or } k = K_{\max}\}$. This exposes state-dependent test-time compute without changing the upstream VLA stack. Using the same lookahead residual in the stopping rule, theory, and threshold scans also keeps the numerical quantity consistent across analysis and deployment.

\textbf{Structural Warm-Start:} Because EqM decoding is independent of a fixed noise schedule, we can exploit temporal coherence between consecutive control cycles. In our implementation, we use a partial warm start: a half-length segment of the previous output chunk is copied into the first half of the current initialization, and the second half is filled with fresh random noise:
\begin{equation}
    A_t^{(0)} = \bigl[ A_{t-1,\mathrm{copy}}^{(\text{out})},\; \epsilon_{\mathrm{tail}} \bigr], \qquad
    A_{t-1,\mathrm{copy}}^{(\text{out})}, \epsilon_{\mathrm{tail}} \in \mathbb{R}^{\frac{H}{2} \times d},
    \label{eq:warmstart}
\end{equation}
where $A_{t-1}^{(\text{out})}$ is the final action chunk returned at control step $t-1$, $A_{t-1,\mathrm{copy}}^{(\text{out})}$ denotes the reused half-chunk from that prediction, and $\epsilon_{\mathrm{tail}}$ is sampled random noise for the remaining half of the horizon. This preserves short-term temporal structure while allowing the remaining horizon to be reinitialized under the new observation. In practice, such warm starts can place the initial guess closer to a local equilibrium basin and reduce the iterations needed to reach a target residual.

\subsection{Theoretical Interpretation of Solver Iterates}
To connect the solver to the stopping rule used later in Sec.~V-C, we provide a local interpretation of early stopping.

\vspace{1ex}
\noindent\textbf{Proposition 1 (Local Potential Descent).}
\textit{Assume that within a neighborhood of the inference trajectory for a fixed $c_t$, the learned field behaves as the gradient of a surrogate energy, $f_\theta(A; c_t) = \nabla E_t(A)$, where $E_t(A)$ is $L$-smooth and lower-bounded by $E_t^\star$. For any step size $\eta \le 1/L$, the iterates generated by Eq.~\eqref{eq:nag} (setting $\mu=0$, so that $\widetilde{A}_t^{(k)} = A_t^{(k)}$) follow a descent path $E_t(A_t^{(k+1)}) \le E_t(A_t^{(k)})$. Using the normalized residual $r_k$ from Eq.~\eqref{eq:residual}, after $K$ steps the minimum residual is bounded by:}
\begin{equation}
    \min_{0 \le j < K} r_j^2 \le \frac{2\bigl(E_t(A_t^{(0)}) - E_t^\star\bigr)}{\eta K H d}.
    \label{eq:bound}
\end{equation}

Moreover, if the solver map is locally contractive with factor $\rho \in (0,1)$ near an equilibrium $A_t^\star$, then
\begin{equation}
    \left\| A_t^{(k)} - A_t^\star \right\|_2 \le \rho^k \left\| A_t^{(0)} - A_t^\star \right\|_2.
    \label{eq:contractive_bound}
\end{equation}
Under the same local $L$-smoothness assumption and the identity $f_\theta(A_t^\star;c_t)=0$, this implies
\begin{equation}
    r_k \le \frac{L}{\sqrt{Hd}} \rho^k \left\| A_t^{(0)} - A_t^\star \right\|_2.
    \label{eq:residual_decay}
\end{equation}
Therefore, a sufficient number of iterations to reach the experimental stopping threshold $r_k \le \tau$ is
\begin{equation}
    K_\tau^{\mathrm{suff}} = \left\lceil \frac{\log\!\left( \frac{L\left\| A_t^{(0)} - A_t^\star \right\|_2}{\tau \sqrt{Hd}} \right)}{\log(1/\rho)} \right\rceil,
    \label{eq:warmstart_iters}
\end{equation}
whenever the logarithm argument exceeds $1$; otherwise the threshold is already met at initialization. Warm-starting reduces this sufficient iteration bound by reducing the initial distance $\left\| A_t^{(0)} - A_t^\star \right\|_2$, up to the ceiling effect in Eq.~\eqref{eq:warmstart_iters}.

This analysis is intentionally local. It justifies the residual threshold used in the experiments as a numerically meaningful stopping rule and explains why warm starts can reduce the iterations needed to reach a target threshold, but it does not determine which threshold maximizes task success. That behavioral question is addressed empirically in Sec.~V-C.

Eq.~\eqref{eq:warmstart_iters} also makes the warm-start effect explicit. Ignoring the ceiling effect, if the initialization distance is reduced by a factor $\alpha \in (0,1)$, then the sufficient iteration count decreases by roughly $\log(1/\alpha)/\log(1/\rho)$. In other words, warm starts are most valuable when the local contraction is strong and the target threshold is strict, which matches the practical intuition that temporal coherence matters most in hard, iterative control regimes.

\section{Experimental Setup}
We evaluate on the 19-task RoboTwin benchmark~\cite{chen2025robotwin}; representative video keyframes are provided in the supplementary material. Each task uses 500 demonstrations (100 clean and 400 randomized), for 9{,}500 total. We compare the original $\pi_0$ against $\pi_0$-EqM while keeping the upstream representation stack and conditioning interface fixed, so only the action expert changes. The main comparison uses a matched 300-step inference budget. Threshold-scan experiments allow up to 1000 EqM iterations and stop when the normalized lookahead residual falls below $\tau$.

We also evaluate on the standard LIBERO suites~\cite{liu2023libero}: LIBERO-Spatial, LIBERO-Object, LIBERO-Goal, and LIBERO-10. For LIBERO, we report suite-level success rates against the corresponding $\pi_0$ baseline.

\section{Results}
\subsection{EqM improves the flow-matching baseline on RoboTwin}
Table~\ref{tab:main19} reports the 19-task deployment results. Under a matched 300-step budget, $\pi_0$-EqM improves 12 tasks, degrades 5, and ties 2, raising the average success rate from 40.4 to 50.2. The largest gains occur on \texttt{pick\_dual\_bottles}, \texttt{place\_cans\_plasticbox}, and \texttt{put\_bottles\_dustbin}.

\begin{table*}[t]
\centering
\small
\caption{Deployment success rates on the 19-task RoboTwin benchmark under a matched 300-step budget. The two methods share the same upstream VLA stack; only the action decoder differs.}
\label{tab:main19}
\begin{tabular}{lccc lccc}
\toprule
Task & $\pi_0$ & $\pi_0$-EqM & $\Delta$ & Task & $\pi_0$ & $\pi_0$-EqM & $\Delta$ \\
\midrule
\texttt{beat\_block\_hammer} & 88 & 85 & $-3$ & \texttt{place\_a2b\_right} & 30 & 30 & $0$ \\
\texttt{click\_bell} & 38 & 33 & $-5$ & \texttt{put\_bottles\_dustbin} & 42 & 72 & $+30$ \\
\texttt{click\_alarmclock} & 24 & 45 & $+21$ & \texttt{stack\_blocks\_two} & 58 & 68 & $+10$ \\
\texttt{handover\_block} & 24 & 27 & $+3$ & \texttt{stack\_bowls\_three} & 68 & 60 & $-8$ \\
\texttt{move\_can\_pot} & 6 & 14 & $+8$ & \texttt{turn\_switch} & 34 & 34 & $0$ \\
\texttt{move\_playingcard\_away} & 66 & 63 & $-3$ & \texttt{pick\_diverse\_bottles} & 20 & 47 & $+27$ \\
\texttt{place\_phone\_stand} & 48 & 55 & $+7$ & \texttt{place\_dual\_shoes} & 54 & 45 & $-9$ \\
\texttt{place\_can\_basket} & 36 & 50 & $+14$ & \texttt{pick\_dual\_bottles} & 26 & 66 & $+40$ \\
\texttt{place\_cans\_plasticbox} & 32 & 65 & $+33$ & \texttt{place\_object\_stand} & 48 & 56 & $+8$ \\
\texttt{place\_a2b\_left} & 26 & 39 & $+13$ & & & & \\
\midrule
Average & 40.4 & 50.2 & $+9.8$ & & & & \\
\bottomrule
\end{tabular}
\end{table*}

\subsection{LIBERO suite results}
Table~\ref{tab:libero} reports the LIBERO results. $\pi_0$-EqM improves from 96.8\% to 97.2\% on LIBERO-Spatial and from 85.2\% to 87.0\% on LIBERO-10, while remaining close on LIBERO-Object and LIBERO-Goal. The four-suite average rises from 94.15\% to 94.35\%. Notably, the strongest gain appears on LIBERO-10, which is consistent with the view that iterative decoding matters most on longer-horizon tasks.

\begin{table}[t]
\centering
\small
\caption{Success rates (\%) on the four LIBERO benchmark suites.}
\label{tab:libero}
\begin{tabular}{lccc}
\toprule
Suite & $\pi_0$ & $\pi_0$-EqM & $\Delta$ \\
\midrule
LIBERO-Spatial & 96.8 & 97.2 & $+0.4$ \\
LIBERO-Object  & 98.8 & 98.4 & $-0.4$ \\
LIBERO-Goal    & 95.8 & 94.8 & $-1.0$ \\
LIBERO-10      & 85.2 & 87.0 & $+1.8$ \\
\midrule
Average & 94.15 & 94.35 & $+0.20$ \\
\bottomrule
\end{tabular}
\end{table}

\subsection{Threshold scans reveal task-dependent non-monotonicity}
\begin{figure}[t]
    \centering
    \includegraphics[width=\linewidth]{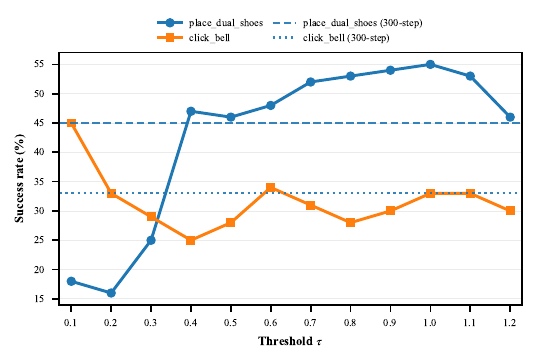}
    \vspace{-2.5em}
    \caption{Threshold scans on two RoboTwin tasks. The preferred threshold differs across tasks, so the same residual threshold changes both solver cost and closed-loop behavior.}
    \label{fig:threshold_scan}
    \vspace{-1em}
\end{figure}

Proposition~1 motivates residual thresholding because it bounds the same normalized residual $r_k$ used in the experiments. Fig.~\ref{fig:threshold_scan} shows scans on \texttt{place\_dual\_shoes} and \texttt{click\_bell}. For \texttt{place\_dual\_shoes}, success peaks near $\tau = 1.0$; for \texttt{click\_bell}, the best result occurs at the strictest tested threshold. The preferred threshold is therefore task dependent.

\begin{figure}[t]
    \centering
    \vspace{-1em}
    \includegraphics[width=\linewidth]{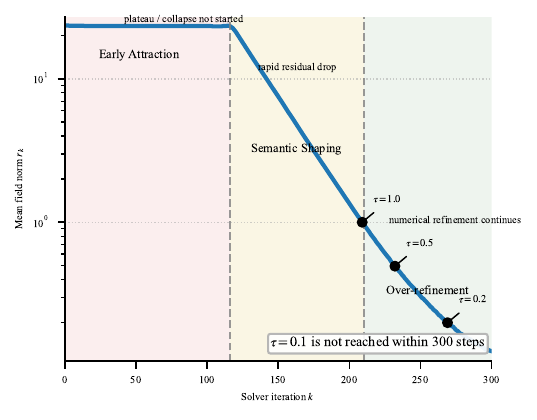}
    \vspace{-2.5em}
    \caption{Qualitative EqM inference trajectory on \texttt{click\_alarmclock}, showing early attraction, semantic shaping, and over-refinement.}
    \label{fig:click_clock_regimes}
    \vspace{-1em}
\end{figure}

Fig.~\ref{fig:click_clock_regimes} gives a single-task view on \texttt{click\_alarmclock}. EqM inference passes through \textit{early attraction}, \textit{semantic shaping}, and \textit{over-refinement}. We use \textit{stationarity--executability gap} for the mismatch between numerical stationarity and physical utility: the iterate with the smallest residual need not execute best.

Together, Proposition~1 and the threshold scans suggest a two-part reading of early stopping: residual thresholds monitor solver progress, but the best threshold remains task dependent. In closed loop, the threshold is therefore not only a numerical tolerance; it also shapes the deployed action distribution.

\section{Discussion and Limitations}
The experiments support three observations: EqM improves RoboTwin under matched compute, transfers competitively to LIBERO, and makes stopping depth behaviorally relevant once decoding becomes iterative. The main limitation is scope: the threshold scans cover only two RoboTwin tasks, so the non-monotonic phenomenon is illustrated rather than mapped exhaustively. Proposition~1 is likewise local: it justifies residual thresholding but does not identify the threshold that maximizes task success or provide a global convergence theorem for the learned non-conservative field. More practically, the current study also does not isolate how much of the gain comes from the EqM objective itself versus the stopping rule and warm-start structure.

At a systems level, the main implication is a cleaner separation between representation learning and deployment-time compute allocation. Once the decoder is time free, one can vary thresholds, iteration caps, or warm-start policies without changing the upstream VLA stack, which turns part of test-time compute into a controllable systems variable rather than a fixed sampler schedule. This perspective may be especially useful in future multi-task and cross-embodiment settings, where the same task-conditioned action field may need to be solved under different latency budgets, action granularities, or robot morphologies. These limitations and opportunities together reinforce the central point that compute allocation in closed-loop VLA should be evaluated behaviorally as well as numerically.

\section{Conclusion}
We presented $\pi_0$-EqM, a decoder replacement for $\pi_0$ that casts action generation as iterative equilibrium solving. Under a matched 300-step budget, it improves RoboTwin average success from 40.4\% to 50.2\% across 19 tasks while remaining competitive on LIBERO and improving LIBERO-10 from 85.2\% to 87.0\%. More broadly, the results argue that inference depth is a policy variable in closed-loop VLA rather than a purely numerical afterthought. They also suggest energy-based VLA---especially composable action generation across tasks and embodiments---as a promising direction for future work.

\bibliographystyle{IEEEtran}
\bibliography{refs}

\end{document}